\documentclass[10pt,twocolumn,letterpaper]{article}

\newcommand\R{\mathbb{R}}

\usepackage{iccv}
\usepackage{graphicx}
\usepackage{amsmath}
\usepackage{amssymb}
\usepackage{booktabs}
\usepackage{graphicx}
\usepackage{amsmath}
\usepackage{amssymb}
\usepackage{booktabs}
\usepackage{times}
\usepackage{epsfig}
\usepackage{graphicx}
\usepackage{amsfonts, amsthm}
\usepackage{comment}
\usepackage{caption}
\usepackage{subcaption}
\usepackage{multirow}
\usepackage{multicol}
\usepackage{enumitem}
 \usepackage{soul}
\newtheorem{thm}{Theorem}[section]

\newtheorem{lemma}[thm]{Lemma}
\newtheorem{prop}[thm]{Proposition}

\newtheorem{defn}{Definition}

\newtheorem{ex}{Example}

\usepackage[pagebackref=true,breaklinks=true,letterpaper=true,colorlinks,bookmarks=false]{hyperref}

\usepackage[capitalize]{cleveref}
\crefname{section}{Sec.}{Secs.}
\Crefname{section}{Section}{Sections}
\Crefname{table}{Table}{Tables}
\crefname{table}{Tab.}{Tabs.}

\iccvfinalcopy % *** Uncomment this line for the final submission

\def\iccvPaperID{11399} % *** Enter the ICCV Paper ID here

\ificcvfinal\fi

\begin{document}

%%%%%%%%% TITLE
%\title{Curvature-Aware Training for\\ Differentiable Coordinate Networks}

\title{Curvature-Aware Training for Coordinate Networks}

\author{Hemanth Saratchandran$^{*1}$
\and
Shin-Fang Chng$^{*1}$
\and
Sameera Ramasinghe$^{2}$
\and
Lachlan MacDonald$^{1}$
\and
Simon Lucey$^{1}$ 
%$^{1}$ School of Computer Science, The University of Adelaide \\
%$^{2}$ Amazon, Australia.
}

\maketitle
\footnotetext[1]{Australian Institute of Machine Learning, University of Adelaide. \\ $^2$ Amazon, Australia. $^*$ Equal Contribution. Correspondence to: Hemanth Saratchandran $<$hemanth.saratchandran@adelaide.edu.au$>$, Shin-Fang Chng $<$shinfang.chng@adelaide.edu.au$>$ }
%\footnotetext[1]{Equal Contribution}
\ificcvfinal\thispagestyle{empty}\fi

%%%%%%%%% ABSTRACT
\begin{abstract}
%Coordinate networks are rapidly becoming key instruments across many computer vision tasks due to their ability to represent signals as compressed, continuous entities. 
%However, training coordinate networks with first-order optimizers typically leads to long training times, hindering their usage in many practical applications. Consequently, recent works that aim for faster training have reverted to differentiable grid-based representations, sacrificing memory for speed. Such works have implicitly highlighted a trade-off between memory efficiency (of coordinate networks) and speed (of grid-based representations). In this work, we theoretically show that by utilizing second-order optimization methods, it is possible to achieve significantly faster training times for coordinate networks while still enjoying their superior compressibility. We then empirically demonstrate the efficacy of our formulation across various signal modalities, including audio, images, videos, signed distance fields, and neural radiance fields.

Coordinate networks are widely used in computer vision due to their ability to represent signals as compressed, continuous entities. However, training these networks with first-order optimizers can be slow, hindering their use in real-time applications. Recent works have opted for shallow voxel-based representations to achieve faster training, but this sacrifices memory efficiency. This work proposes a solution that leverages second-order optimization methods to significantly reduce training times for coordinate networks while maintaining their compressibility. Experiments demonstrate the effectiveness of this approach on various signal modalities, such as audio, images, videos, shape reconstruction, and neural radiance fields.
\end{abstract}

%%%%%%%%% BODY TEXT
\vspace{-0.5cm}

\section{Introduction}
\label{sec:intro}

Coordinate networks \cite{sun2021coil}, or implicit neural functions \cite{siren}, achieve state-of-the-art results in multidimensional signal reconstruction tasks, such as image synthesis \cite{skorokhodov2021adversarial, chen2021learning}, geometry \cite{sitzmann2019scene, nerf}, and robotics \cite{li20223d, chen2022fully}. However, coordinate networks admitting traditional activation functions (e.g., ReLU, sigmoid, and tanh) fail to capture high-frequency details due to spectral bias \cite{rahaman2019spectral}. To overcome this limitation, positional embedding layers \cite{tancik2020fourier} are often added, but they can produce noisy first-order gradients that hinder architectures requiring backpropagation \cite{lin2021barf, chng2022garf}. A recent alternative approach is to use non-traditional activation functions, such as sine \cite{siren} or Gaussian \cite{ramasinghe22} activations, which enable high-frequency encoding without positional embedding layers. The major benefit of these activations over positional embedding layers is their well-behaved gradients~\cite{siren,ramasinghe22}.

%Despite demonstrating remarkable performance in signal reconstruction tasks, coordinate networks are typically trained with first-order optimizers such as Adam, resulting in slow training times. As a result, certain members of the vision community have reverted to using grid-based representations \cite{fridovich2022plenoxels, trading_pos}, despite their drawbacks, such as high memory usage and lack of inherent architectural bias.

Although coordinate networks have shown remarkable performance in signal reconstruction tasks, they are typically trained using first-order optimizers like Adam, leading to slow training times. Consequently, some in the vision community 
have resorted to using shallow voxel-based 
representations~\cite{fridovich2022plenoxels, trading_pos, chen2022tensorf,sun2022direct}, despite their drawbacks such as high memory usage and lack of implicit architectural bias.

\begin{figure}
    \centering
     \includegraphics[width=1.0\linewidth]{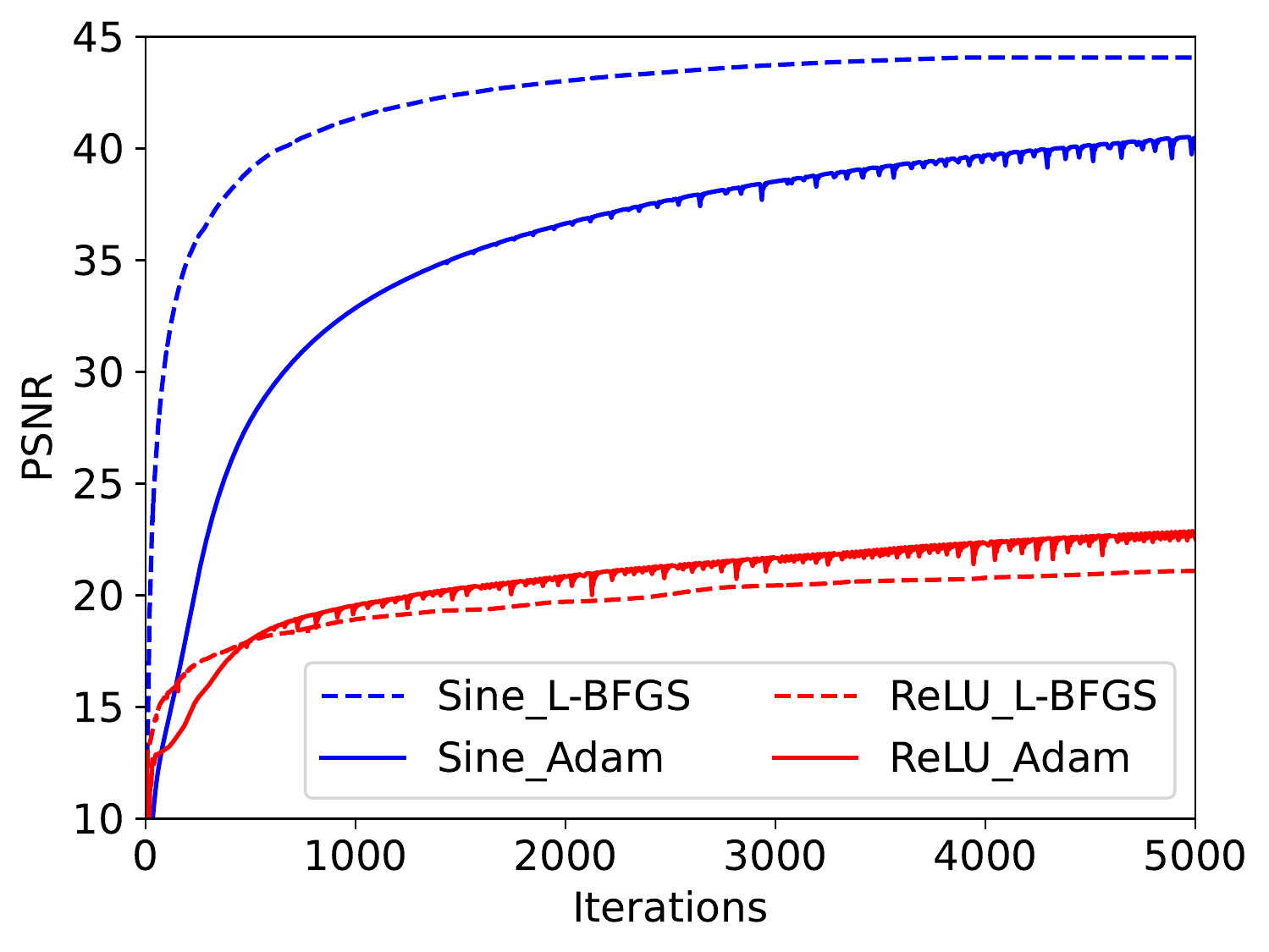}
        \vspace{-2em}
    \caption{Sine- and ReLU-coordinate networks were compared on an image reconstruction task using L-BFGS and Adam optimizers. The L-BFGS optimizer showed faster convergence for the sine-network, while the ReLU-network converged faster with Adam.}     \label{fig:fig_1}
\end{figure}
%\begin{figure}
%    \centering
%     \includegraphics[width=1.0\linewidth]{figures/image/fig1_final_test.pdf}
%     \hspace{-0.2cm}
%    \caption{sine vs ReLU-activated coordinate network when trained with L-BFGS. \textit{Top:} PSNR convergence. \textit{Bottom}: Reconstruction of sine activated (\emph{left}) and ReLU-activated (\emph{right}) after training on 1 million points for 1000 iterations.}     \label{fig:fig_1}
%    \vspace{-1em}
%\end{figure}

%Comparison of training convergence for coordinate networks activated with sine and ReLU functions, trained with L-BFGS and Adam optimizers, on an image reconstruction task.
%The rate of convergence for the L-BFGS optimizer on a sine-network is significantly faster than Adam. However, for the ReLU-network the opposite occurs.

%In parallel, another recent line of research has focused on investigating positional-embedding-free coordinate-networks. At the heart of these methods lie non-traditional activation functions, that allow the networks to encode high-frequency details without positional embedding layers. Two prime examples of such activations are sine \cite{siren} and Gaussian \cite{ramasinghe22} activations. The major benefit of these activations over positional embedding layers is their well-behaved gradients and curvature.

%\hl{It is well understood empirically}~\cite{?} that nearly all coordinate/implicit neural networks are trained using an Adam (or variant) optimizer due to its rapid convergence and high signal fidelity. 

In this paper, we present an intriguing revelation that a new breed of coordinate networks~\cite{siren, ramasinghe22}, activated by 
sine and Gaussian functions, can be trained efficiently using second-order optimizers such as L-BFGS \cite{nocedal1999numerical}. This is because their loss landscapes exhibit favorable gradient and curvature conditioning, which leads to superlinear convergence, in contrast to the linear convergence seen with Adam. Fig. \ref{fig:fig_1} showcases this point by comparing sine- and ReLU-coordinate networks trained with an L-BFGS optimizer \cite{nocedal1999numerical}, a curvature-aware second-order optimizer, and an Adam optimizer. The convergence rate of the sine-network trained with L-BFGS is significantly faster than the one trained with Adam -- highlighting the good curvature properties of the loss landscape. In contrast, the ReLU-network trained with Adam has faster convergence rate than the one trained with L-BFGS, a manifestation of the poor curvature properties of its loss landscape.

However, one of the downsides of second-order optimizers is their computational complexity when dealing with a large number of parameters. We explore this issue in the context of coordinate networks and demonstrate that, as the network size grows, Adam may outperform L-BFGS in terms of training time. To address this challenge, we propose a novel strategy of breaking down large-scale datasets into smaller patches and training a single coordinate network with a second-order optimizer on each patch. Our experiments reveal that this approach can lead to training time accelerations of up to $14$ times faster than Adam, and serves as an effective remedy for modelling larger size signals.

A summary of our contributions are:-
\begin{enumerate}
    
    \item Our paper is the first to examine the training of coordinate networks using L-BFGS and show theoretically that while superlinear convergence is guaranteed for 
    networks activated by sine or Gaussian functions,
    it is not generally guaranteed for ReLU (with or without positional embedding).
    \item We validate this theory empirically by showing that sine-/Gaussian-activated coordinate network's are up to $5$ times faster when trained with L-BFGS over Adam. We present results on image, audio, video, shape and radiance field reconstruction tasks. 
    \item A patch-based decomposition strategy is explored to limit the considerable computational cost of L-BFGS as the size of the signal/network grows. Specifically, we demonstrate that a sine-activated patch-based NeRF (i.e. KiloNeRF~\cite{kilonerf}) trained with L-BFGS is $6$ times more efficient than the same network trained with Adam.

    %Specifically, we demonstrate this strategy on images and NeRF by showing that a sine-activated patch-based coordinate-network trains with L-BFGS upto 14 times faster than Adam on such instan 
    %this strategy on images and NeRF showing that

    %Specifically, we demonstrate that strategy on images and NeRF (~i.e. KiloNeRF\cite{kilonerf}) with L-BFGS up to $14 \times$ more efficient than the same network trained with Adam.
    %Specifically, we demonstrate that a sine-activated patch-based NeRF (i.e. KiloNeRF~\cite{kilonerf}) with L-BFGS is six-times more efficient than the same network trained with Adam 

     % \item An in-depth comparative analysis of the computational complexity of various second-order optimizers for coordinate networks.
\end{enumerate}

\section{Related Work}
\label{sec:relwork}

%\textbf{Implicit neural representations:} Several recent works have shown the high powered 
%performance of implicitly defined continuous signal representations parameterised by 
%neural networks, colloquially known as coordinate-MLPs, over conventional grid based 
%approximations (references). One of the earliest works in this area 
%came from Mildenhall et al. (reference) where such networks were used to give high 
%fidelity reconstructions of photometric veiw projections from an arbitrary angle. This 
%ground breaking work led to several applications of neural implicit representations in 
%areas such as shape reconstruction (reference), 3D object reconstruction (reference), 
%and novel view synthesis (reference). For optimal performance, such MLPs had to be precomposed with a positional embedding layer so as to be able to learn high frequency content in the target signal. 

\paragraph{Coordinate Networks \cite{sun2021coil}} also known as implicit neural functions \cite{siren}, have gained increasing interest in recent years due to the seminal work by Mildenhall et al.~\cite{nerf}. Unlike conventional neural networks that operate on high-dimensional inputs and are primarily used for classification tasks, coordinate networks encode signals as weights using low-dimensional coordinates and aim to preserve smoothness in the outputs \cite{trading_pos}. One of the remarkable aspects of Mildenhall et al.'s work is their demonstration of the generalization properties of neural signal representations, which ushered in a huge body of work on the subject in recent years \cite{chen2019learning, Deng20, Genova_2020_CVPR, mulayoff2021implicit, Park_2019_CVPR, park2021nerfies, pumarola2021d, Rebain_2021_CVPR, Saito_2019_ICCV, siren, chng2022garf, wang2021nerf, Yu_2021_CVPR, saragadam2022miner,zhu2022nice,chen2023factor}. However, to achieve optimal performance, such networks need to use positional embeddings to encode high-frequency signal content \cite{trading_pos}. Sitzmann et al. \cite{siren} proposed SIREN, a sine-activated network, that can improve the fidelity of signals without positional embedding layers. However, a disadvantage of SIREN is that it needs a principled initialization scheme \cite{siren}. 
Ramasinghe and Lucey \cite{ramasinghe22} introduced a Gaussian activated coordinate network that, like SIREN, achieved state-of-the-art performance on signal reconstruction but is robust to random initialization schemes.

\vspace{-1em}
%\paragraph{Optimisaton of Neural Networks} is a vast subject with a deep history. Classically, practitioners in the field trained neural networks with gradient decent, primarily for its ease of use and low memory requirements. As more sophisticated neural network models took centre stage, gradient decent saw many variants that allowed practitioners to use large models without incurring huge memory footprints. In the setting of theoretical optimization, second-order optimization methods are a very powerful group of algorithms that admit superior convergence when compared to first order ones \cite{nocedal1999numerical}. However, a major drawback in using such algorithms in practise is their large computational cost \cite{nocedal1999numerical}. A further drawback is that such algorithms are in general not amenable to the use of stochastic sampling strategies, a method that has had great success when applied to first order gradient decent methods. This has led researchers to develop various second-order optimizers for machine learning applications that do not use such a high computational cost, as is the case with classical second-order algorithms \cite{nocedal1999numerical}. Algorithms such as K-FAC \cite{martens2015optimizing}, variants of L-BFGS
%\cite{boggs2019adaptive, zhao2017stochastic, moritz2016linearly}, Shampoo \cite{gupta2018shampoo}, and GGT \cite{agarwal2019efficient} are showing that it is possible to train neural networks with second-order methods, yielding superior results over standard first order ones.

\paragraph{Optimizaton of Neural Networks}
is a complex topic with a rich history. Initially, practitioners used gradient descent due to its ease of use and low memory requirements. However, as more sophisticated models emerged, variants of gradient descent were developed to accommodate larger models. While second-order optimization methods offer superior convergence in theory \cite{nocedal1999numerical}, they are computationally expensive and not easily applicable to stochastic sampling strategies. To overcome these challenges, researchers have developed second-order optimizers that yield superior results compared to standard first-order ones, such as K-FAC \cite{martens2015optimizing}, variants of L-BFGS \cite{boggs2019adaptive, zhao2017stochastic, moritz2016linearly}, 
Shampoo \cite{gupta2018shampoo}, and GGT \cite{agarwal2019efficient}.

%\textbf{Neural scene representations:} A great accomplishment of coordinate networks has been in their 
%application to synthesising novel views of a complex scene. Of particular importance is NeRF 
%\cite{nerf}, a ReLU based coordinate-MLP, with positional embedding, that synthesises a novel view of a complex scene using a volume rendering framework to fit a neural radiance field to an RGB image. 
%Since this ground breaking work there have been several applications of coordinate-MLPs to a variety of areas in novel synthesis and shape reconstruction 
%\cite{Chen_2019_CVPR, Deng20, Tiwari_2021_ICCV, Genova_2020_CVPR, Park_2019_CVPR, Niemeyer_2020_CVPR, Saito_2019_ICCV, Yu_2021_CVPR, Rebain_2021_CVPR, park2020deformable}.
%A disadvantge of NeRF is its slow rendering times. KiloNeRF \cite{kilonerf} is a new innovative architecture 
%that tackles the slow rendering times of NeRF. Utilising several tiny MLPs to represent a tiny portion 
%of a complex scene, and using a teacher-student distillation, KiloNeRF achieves three orders of 
%magnitude faster rendering times than NeRF. 
%Non-traditional activations have also found applications in architectures 
%synthesising novel views of a complex scene, with the introduction of Gaussian activated neural radiance fields (GARF) \cite{garf}, showcasing state of the art performance in terms of high fidelity reconstruction and pose estimation.

%\textbf{Optimisation:} 

\section{Preliminaries}
\subsection{Coordinate Multi-Layer Perceptrons (MLPs)}\label{subsec:coordinate_nets_prelims}
Coordinate-MLPs are a new class of neural networks which encode signals 
as weights using low dimensional coordinates as inputs.
A coordinate-MLP with $L$ layers, 
$f : \R^{n_0} \rightarrow \R^{n_L}$ can be defined as
\begin{equation}\label{NN}
  f : x \rightarrow T_L \circ \psi \circ T_{L-1} \circ \cdots,
\circ \psi \circ T_L(x),  
\end{equation}
where $T_i : x_i \rightarrow A_ix_i + b_i$ is an affine transformation with trainable parameters $A_i \in \mathbb{R}^{n_{i-1}\times n_i}$, $b_i \in \mathbb{R}^{i}$, and $\psi$ is a non-linear activation. The layer widths of the network are given by the numbers $\{n_1, n_2, \ldots, n_L \}$. 
%The coordinate networks we consider will all be trained with the MSE loss function as the objective function.

All our networks will be trained with the Mean Squared Error (MSE) loss function.
Given  $N$ training samples 
$\{(x_i, y_i)\}_{i=1}^N$, where $x_i$ and $y_i$ denotes the input data and target data, respectively, we write the MSE loss function as
\begin{equation}\label{loss}
   \mathcal{L}(\theta) = \frac{1}{N}\sum_{i=1}^N\frac{1}{2}|f(\theta, x_i) - y_i|^2,  
\end{equation}
where $f$ denotes a coordinate-MLP, and $\theta$ denotes the parameters of $f$, i.e. the weights and biases $(W,b)$. 

We briefly discuss commonly used coordinate-MLPs. 
\vspace{-1em}
\paragraph{ReLU-MLPs} are popular due to their universal approximation capabilities \cite{huang2020relu, liu2021optimal}, but they suffer from spectral bias \cite{rahaman2019spectral}. This bias can cause a preference for low-frequency components, leading to suboptimal signal reconstruction, particularly for signals with high-frequency components.

%\paragraph{ReLU MLPs} which employ 
%$ReLU(x) = \max(0,x)$, have been widely used due to their ability to be universal approximators. However, a major drawback of ReLU MLPs is their tendency to exhibit spectral bias~\cite{rahaman2019spectral}, which can lead to biased learning of low-frequency components of a signal. This can result in suboptimal reconstruction of signals that contain high-frequency components.
%A major disadvantage with such networks is that they admit spectral bias \cite{rahaman2019spectral}, causing them to bias the learning of low frequency components of a signal, leading to sub-optimal reconstruction of signals with high frequency components. 
\vspace{-1em}
\paragraph{Positional encoded MLPs (ReLU-PE)} avoid the spectral bias of ReLU-MLPs by adding a positional embedding layer (\textbf{PE}) to the network. This involves embedding low-dimensional data inputs \textbf{x} into a higher-dimensional space using an embedding layer $\gamma : \R^d \rightarrow \R^{d + D}$. Popular embedding layers include Fourier feature embeddings \cite{tancik2020fourier} and Gaussian embeddings \cite{trading_pos}.

%The resulting embedded data is then passed through a standard ReLU MLP, creating a neural network $F(\theta, \gamma(\textbf{x}))$. Popular embedding layers include Fourier feature embeddings \cite{tancik2020fourier} and Gaussian embeddings \cite{trading_pos}.

%\paragraph{Positional encoded MLPs (ReLU-PE)} circumvent the spectral bias of ReLU MLPs by adding a positional embedding (\textbf{PE}) layer to the network. Low dimensional data inputs \textbf{x} are embedded into a high dimensional space via an embedding layer $\gamma : \R^d \rightarrow \R^{d + D}$. The resulting embedded data points are then passed through a standard ReLU MLP, obtaining a neural network 
%$F(\theta, \gamma(\textbf{x}))$, where $\theta$ denotes the parameters and 
%$\textbf{x}$ the data. Some of the
%most commonly used embedding layers are 
%Fourier feature embeddings \cite{tancik2020fourier} and Gaussian embeddings \cite{trading_pos}.
\vspace{-1em}
\paragraph{Sine-MLPs} are a positional embedding free coordinate network~\cite{siren} that employ a sine activation function
$\textbf{x} \rightarrow \sin(2\pi\omega\textbf{x})$, where $\omega$ is a frequency hyperparameter. A larger $\omega$ increases the frequency of the network allowing it to learn high-frequency targets, overcoming spectral bias.
\vspace{-2em}
\paragraph{Gaussian-MLPs} are another class of embedding-free coordinate networks~\cite{ramasinghe22} that employ a Gaussian activation function 
$\textbf{x} 
\rightarrow \exp\big{(} \frac{|\textbf{x} - \mu|^2}{2\sigma^2} \big{)}$. The hyperparameter
$\mu$ denotes the mean and $\sigma$ the standard deviation of the Gaussian, with a smaller 
$\sigma$ leading to a higher frequency network.

\subsection{Second-order optimizers}\label{subsec:second_order_optim}
%The coordinate networks we consider will all be trained with the MSE loss function as the %objective function. Given a collection of $N$ training samples 
%$\{(x_i, y_i)\}_{i=1}^N$, where $x_i$ denote the input data and $y_i$ target data, we %write the MSE loss function as:
%\begin{equation}\label{loss}
%   \mathcal{L}(\theta) = \frac{1}{N}\sum_{i=1}^N\frac{1}{2}|F(\theta, x_i) - y_i|^2  
%\end{equation}
%where $F$ denotes a neural network function, and $\theta$ denotes the parameters of $F$, %i.e. the weights and biases $(W,b)$. 

%This section provides a brief overview of three main second-order optimizers that will be discussed throughout the paper. For a more detailed analysis, including pseudocode, of these and other second-order optimizers, we refer readers to the supp. material. %is given in the supp. material. The reader who is unfamiliar with second-order optimization algorithms is kindly asked to consult the supp. material.

This section introduces 
three second-order optimizers discussed in the paper. For more information and pseudocode, see Sec. 1 of supp. material.
\vspace{-1em}
\paragraph{Newtons method}utilizes a quadratic approximation of an objective function $f$ and uses the inverse Hessian matrix to take a gradient step. The update is computed as:
\begin{equation}\label{newton update}
    \theta_{t+1} = \theta_t - H(\theta_t)^{-1}\nabla f(\theta_t),
\end{equation}
where $H(\theta_t)$ denotes the Hessian of $f$ at $\theta_t$. Thus we see that
the optimizer utilizes curvature information to take updates as the Hessian is a measure of the curvature of the objective function.
Convergence to a global minimum is guaranteed for convex functions~\cite{nocedal1999numerical},
but the algorithm may not converge for non-convex functions. When the algorithm converges to a minimum, it does so at a \textit{quadratic rate}, which is significantly faster than first-order optimizers such as gradient descent/Adam, which converge at sub-linear/linear rate \cite{nocedal1999numerical}.
However, inverting the Hessian has a computational complexity of $\mathcal{O}(p^3)$ for an objective function with $p$ parameters~\cite{nocedal1999numerical}, making Newton's method  memory-intensive for high parameter objective functions such as overparameterised neural networks.

\vspace{-1em}
\paragraph{The BFGS algorithm} 
is a quasi-Newton method that approximates the inverse Hessian matrix with a positive definite matrix $M_t$ iteratively to avoid computing the exact Hessian matrix in Newton's method.

%seeks to achieve the advantages of Newton’s method without the computational complexity of computing the exact Hessian. Such methods are known as Quasi-Newton methods. The algorithm
%iteratively approximates the inverse Hessian $H^{-1}$, in \eqref{newton update}, with a positive definite matrix $M_t$. 
Given a choice of initialisation, 
$M_0$, $M_{t+1}$ can be computed using the closed form BFGS update
\begin{equation}\label{bfgs update}
    M_{t+1} = \bigg{(}I - \frac{y_ts_t^T}{y_t^Ts_t}\bigg{)}^TM_t
    \bigg{(}I - \frac{y_ts_t^T}{y_t^Ts_t}\bigg{)} + \frac{s_ts_t^T}{y_t^Ts_t},
\end{equation}
where for an objective function $f$
\begin{equation}\label{curvature_pair1}
    y_t = \nabla f(\theta_{t+1}) - \nabla f(\theta_t) \hspace{0.1cm}\text{ and } \hspace{0.1cm}
     s_t = \theta_{t+1} - \theta_t.
    \end{equation}
%and
%\begin{equation}\label{curvature_pair2}
%    s_t = \theta_{t+1} - \theta_t.
%\end{equation}
The parameter update at iteration $t$ is then given by
\begin{equation}\label{param_update_bfgs}
    \theta_{t+1} = \theta_t - M_t\nabla f(\theta_t).
\end{equation}
The fundamental principle of quasi-Newton methods is to avoid computing the inverse Hessian from scratch every iteration. Instead, the BFGS algorithm approximates the inverse Hessian $H^{-1}$ with a positive definite matrix $M_{t+1}$, via \eqref{bfgs update}, using recent curvature information, 
via \eqref{curvature_pair1}, 
and an existing approximation $M_t$. This reduces the computational complexity to $\mathcal{O}(p^2)$ for an objective function with $p$-parameters. The BFGS algorithm converges at a superlinear rate, slower than Newton's method but faster than Gradient descent/Adam, for a twice differentiable objective function with Lipshitz continuous Hessian \cite{nocedal1999numerical}.

\vspace{-1em}
\paragraph{The L-BFGS algorithm} is a limited memory variant of the BFGS algorithm.
Instead of storing the approximate inverse Hessian $M_t$ at each iteration, the algorithm
stores a limited number of the vector pairs $\{s_t, y_t\}$, see~\eqref{curvature_pair1}, used in the construction of the approximate Hessian $M_t$, see
\eqref{bfgs update}. This reduces the computational complexity to 
$\mathcal{O}(p)$ \cite{nocedal1999numerical}. Its convergence rate is superlinear \cite{nocedal1999numerical}.

%Using a second-order optimizer can significantly improve the convergence rate of neural network training compared to traditional optimizers like SGD and Adam, which generally have sub-linear/linear convergence rates \cite{garrigos2023handbook, kingma2014adam}. However, memory complexity can be a serious issue for second-order optimizers. We address this issue in sec.~\ref{sec:kilo_image} and \ref{sec:nerf}.

%reducing the computational complexity to $\mathcal{O}(p)$ \cite{nocedal1999numerical}. 

%Our experiments will be run using an L-BFGS optimizer due to its computational complexity %being much less then various other second-order algorithms. Furthermore, 
%standard machine learning software such as PyTorch, TensorFlow and JAX all support an 
%L-BFGS implementation. 

%In the supp. material, we make comparisons of L-BFGS with other second-order %optimizers such as Conjugate Gradient and K-FAC, highlighting the low memory footprint of %L-BFGS.

\begin{figure*}[t]
    \centering
    \begin{subfigure}[t]{0.5\textwidth}
    \includegraphics[width=\textwidth]{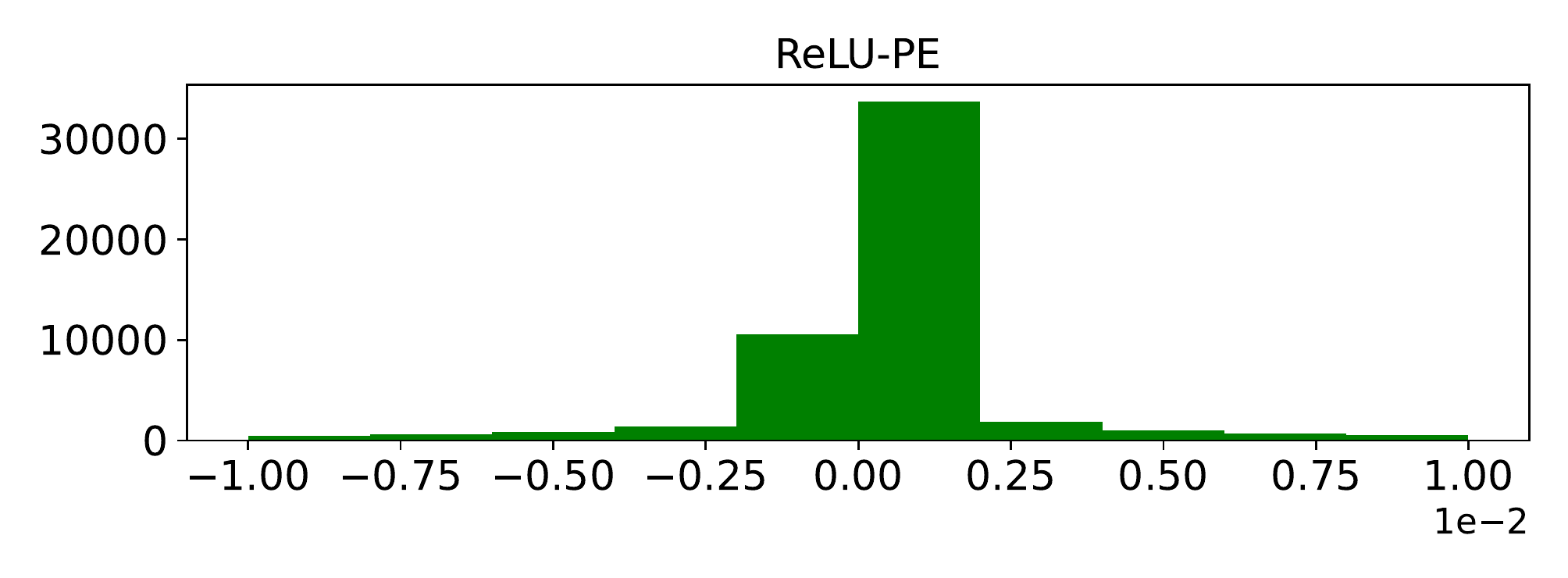} 
    %\caption{Distribution of Hessian eigenvalues of MSE loss after 10 iterations}
    \label{fig:eigenvals1}
    \end{subfigure}
    \begin{subfigure}[t]{0.45\textwidth}
    \includegraphics[width=\textwidth]{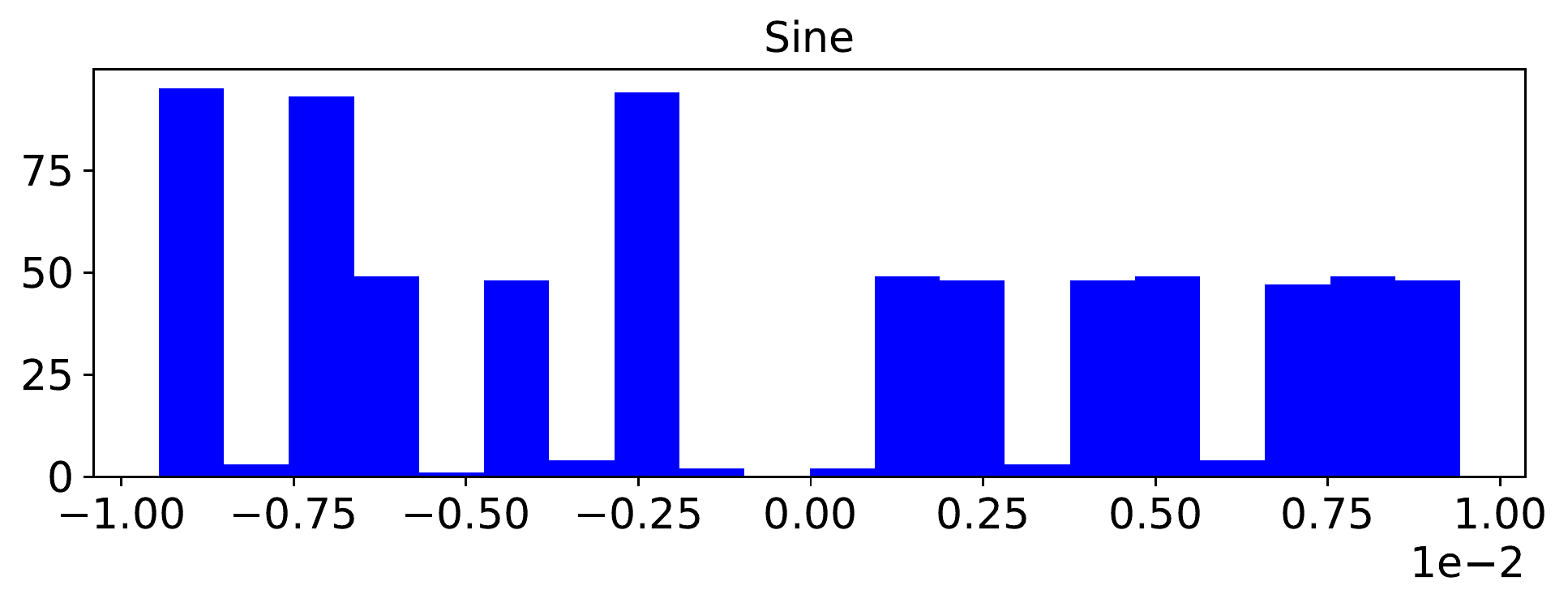}
    %\caption{Distribution of Hessian eigenvalues of MSE loss after 50 iterations}
    \label{fig:eigenvals2}
    \end{subfigure}
    \label{fig:eigs}
    \vspace{-2.5em}
    \caption{Total eigenvalue distribution of the Hessian of MSE loss for sine- and ReLU-PE-activated networks throughout training. ReLU-PE has $28\%$ of its eigenvalues at $0$, while the smallest eigenvalue for the sine-activated network is $5\times 10^{-4}$. This highlights the superior conditioning of the Hessian of a sine-activated network (\textbf{no} zero eigenvalues) compared to a ReLU one (\textbf{many} zero eigenvalues).}    \label{fig:eigenval_training}
  \end{figure*}

%\section{Methodology}
\section{Theoretical Analysis}\label{sec:hess_theory}

%\subsection{Hessian analysis: A theoretical approach}\label{hess_theory}

\subsection{Analyzing the Hessian of a Coordinate Network}\label{hess_theory}
This section gives a theoretical analysis of the poor gradient and curvature conditioning of the MSE loss landscape of a ReLU-activated coordinate network. In contrast, the well-conditioned gradient and curvature of the MSE loss landscape of a sine-/Gaussian-coordinte network is highlighted. The predictions made from the theory are then empirically verified.

%sinusoidal-/Gaussian-activated coordinate network are highlighted. The predictions made from the theory are then theoretically verified.

%This section provides a theoretical analysis of the Hessian of the MSE loss, see \eqref{loss}, of a bias-free coordinate-MLP. The supp. material contains the proofs, which also apply to networks with biases. 

%In this section, we give a theoretical analysis of the Hessian of the MSE loss of a coordinate network. We will assume the networks in consideration have no biases, so as to make the analysis easier. Furthermore, we will denote the MSE loss of our networks by
%$\mathcal{L}$.
%All the results of this section go through for networks with biases, as shown in the supp. material.
%The details and proofs of the results of this section can be found in the supp. material.

As the weights of a neural network in \eqref{NN}  
are trainable we can represent it as a map 
$f : \R^p \times \R^d \rightarrow \R^{n_L}$, where $p$ denotes the parameter dimension and is given by $p = n_0\times n_1 + n_1\times n_2 + \ldots + n_{L-1} \times n_L$. 
Letting 
$p = (\theta_1,\ldots,\theta_L)$ with $\theta_i \in \R^{n_i\times n_{i-1}}$, we write the map as
\begin{equation}\label{NN_as_comp}
    f(\theta, X) = f_{L}(\theta_L)\circ\cdots\circ f_{1},(\theta_{1})(X),
\end{equation}
where $f_{i}(\theta_i) : \R^{n_{i-1}} \rightarrow \R^{n_i}$ is defined by 
\begin{equation}
 f_{i}(\theta_i)(v) = \psi(\theta_i\cdot v).   
\end{equation}
Each of the maps $f_{i}(\theta_i)$ can be expanded as a map
\begin{equation}
    f_i :  \R^{n_i\times n_{i-1}} \times  \R^{n_{i-1}} \rightarrow \R^{n_i}
\end{equation}
and thus a neural network can equally well be described via a collection of maps
$\{f_i :  \R^{n_i\times n_{i-1}} \times  \R^{n_{i-1}} \rightarrow \R^{n_i}\}_{i=1}^{n_L}$ satisfying the composition structure \eqref{NN}.

Given an input data
set $X$, we let
\begin{equation}
    F_k = f_{k}(\theta_k)\circ\cdots\circ f_1(\theta_1)(X)
\end{equation}
denote the $k$-layer neural output function.

For a fixed set of training data $(X,Y)$, with $X \in \R^d$ the inputs and $Y \in \R^{n_L}$ the targets, the MSE loss function, see \eqref{loss}, is a map $\mathcal{L} : \R^p \rightarrow \R$. To simplify the statement of the following lemma, we introduce the following notation. Let $\Delta(\psi'(\theta_{L-j}F_{L-j-1})$ denote the diagonal matrix with entries given by
$\psi'(\theta_{L-j}F_{L-j-1})$, where $\psi'$ denotes the derivative of the activation function $\psi$, flattened column wise as a vector and let
\begin{equation}
    \mathcal{D}_{L-l-1} = 
\prod_{j=1}^{L-l-1}(\theta_{L-j}^T\otimes Id)\Delta(\psi'(\theta_{L-j}F_{L-j-1})),
\end{equation}
where $\otimes$ denotes the Kronecker product of matrices.

%Using equation \eqref{loss} and the chain rule we have the following lemma that computes the %gradient of the loss.
The following lemma computes the gradient of the MSE loss \eqref{loss}.
\begin{lemma}\label{grad of loss}
Let $f$ be a neural network and $(X, Y)$ a training data set, with $X$ inputs and $Y$ targets, defined by the family of maps 
$\{f_i :  \R^{n_i\times n_{i-1}} \times  \R^{n_{i-1}} \rightarrow \R^{n_i}\}_{i=1}^L$. Given 
$\theta_l \in \R^{n_l \times n_{l-1}}$, we have
\begin{equation}\label{grad_loss_form}
    \nabla_{\theta_l}\mathcal{L} = 
    \big{(}Id\otimes F_{l+1}\big{)}\mathcal{D}_{L-l-1}\big{(}\theta_L^T\otimes Id \big{)}
    \big{(}F_L - y \big{)},
\end{equation}
where $\mathcal{L}$ denotes the MSE loss function associated to the network.
\end{lemma}
The Hessian of the MSE loss of a neural network can be computed using lemma \ref{grad of loss}, the product rule, and the chain rule. 
%Given a composition of functions $f \circ g$, we have the formula
%$Hess(f\circ g)(z) = Jg(z)^T\cdot Hess(f)(g(z))\cdot Jg(z) + Jf(g(z))\cdot H(f)(z)$. Applying this %formula together with \eqref{grad_loss_form} allows one to obtain a formula for the Hessian of the %loss.
For each point $\theta \in \R^p$, the Hessian $Hess(\mathcal{L}(\theta))$ will be a 
$(n_L\times p) \times p$-matrix.
Thus one can see that if $p$ is large, the Hessian will be an extremely large matrix even in the case that $n_L = 1$. Even though the Hessian is an enormous matrix, one can still obtain insight into its structure via \eqref{grad_loss_form}. 
Given a parameter point $\theta_k \in \R^{n_k \times n_{k-1}}$, we observe that the second derivative $\nabla_{\theta_k}\nabla_{\theta_l}\mathcal{L}$
%\begin{equation}
%      \nabla_{\theta_k}\nabla_{\theta_l}\mathcal{L}
%\end{equation}
will have three main terms given by applying the product rule:
\begin{itemize}
    \item[1.] $\big{(}\nabla_{\theta_k}\big{(}Id\otimes F_{l+1}\big{)}\big{)}\mathcal{D}_{L-l-1}\big{(}\theta_L^T\otimes Id \big{)}
    \big{(}F_L - y \big{)}$
    
    \item[2.] $\big{(}Id\otimes F_{l+1}\big{)}
    \big{(} \nabla_{\theta_k}\mathcal{D}_{L-l-1}\big{)}\big{(}\theta_L^T\otimes Id \big{)}
    \big{(}F_L - y \big{)}$
    
    \item[3.] $\big{(}Id\otimes F_{l+1}\big{)}\mathcal{D}_{L-l-1}
    \big{(}\nabla_{\theta_k}\big{(}\big{(}\theta_L^T\otimes Id \big{)}
    \big{(}F_L - y \big{)}\big{)}\big{)}$.
\end{itemize}
Terms 1. and 3. will all contain first-order derivatives of the neural network function $f$. The second term will be the only term that will contain second-order derivatives of the neural network function, see Sec. 2 of supp. material for details.
As the derivative of a ReLU activation is a step function, and its second derivative is a Dirac delta distribution, see Sec. 2 of supp. material for a proof,
this analysis shows that in the case of a ReLU-activated network (with or without positional embedding), the Hessian of the loss function is discontinuous and hence poorly conditioned.

\begin{prop}\label{hessian relu}
Let $f$ be a ReLU-network, with or without positional embedding. Then the hessian of the loss
$\mathcal{L}$ contains two types of poorly conditioned terms:
\begin{itemize}
    \item[1.] sums of step functions
    \item[2.] sums of Dirac delta distributions.
\end{itemize}
\end{prop}
%The above proposition shows that the Hessian of the loss of a ReLU network, with or without  positional embedding, will be rank deficient due to the potential of a large number of zeros in the Hessian matrix coming from the step function and delta function terms in the Hessian matrix expansion. We verify this insight experimentally in section \ref{hess empirical}. 
%In contrast, the derivatives of a sine function is another sine function (a cosine is a sine with a phase offset) and the derivatives of a Gaussian function are just sums of Gaussians multiplied by polynomials. Such derivatives are far better behaved than a step function or a Dirac delta distribution. We show experimentally in section \ref{hess empirical} that the hessian of such coordinate networks in general do not have any zero eigenvalues and hence are not rank deficient. 

Prop.~\ref{hessian relu} implies the Hessian of the loss function in a 
ReLU-network, with or without positional embedding, is likely to be rank deficient due to the high probability of encountering many zeros in the Hessian matrix arising from the step function and delta function terms in its expansion. 
In contrast, the derivatives of sine and Gaussian functions
exhibit smoother behavior and are less prone to rank-deficiency in their Hessians. Additionally, Prop.~\ref{hessian relu} suggests that the curvature of a ReLU-MLP is poorly conditioned compared to one activated by sine/Gaussian.
Hence, second-order optimizers that take curvature into account are expected to perform better on sine- or Gaussian-activated coordinate networks as opposed to those activated by ReLU.

We verified our theoretical predictions on an image reconstruction task by training two networks: one with a sine activation and another with a ReLU-PE~\cite{nerf}. Both networks were trained for $50$ iterations on a $50\times50$ image with full sampling and L-BFGS optimizer. We computed the eigenvalues of the Hessian of the MSE loss at each iteration throughout training. Fig. \ref{fig:eigenval_training} shows the distribution of eigenvalues in the interval $[-0.01, 0.01]$. As predicted by our theory, the sine-activated network has no zero eigenvalues, whereas the ReLU-PE-network has many. For a more comprehensive analysis, including ReLU and Gaussian MLPs with varying width, depth, and initialization schemes, refer to Sec. 3 of supp. material.

\subsection{Analyzing L-BFGS on a Coordinate Network}
In this section, we provide a theoretical and empirical analysis of the 
L-BFGS  algorithm \cite{nocedal1999numerical} on coordinate networks activated by ReLU and sine/Gaussian.
We theoretically show that for a 
ReLU/ReLU-PE-activated coordinate network the L-BFGS algorithm is not guaranteed to converge superlinearly, however for a sine- or Gaussian-activated
network superlinear convergence is guaranteed. We then verify these theoretical predictions empirically.

%sinusoidal/Gaussian network it is. We then verify our theoretical predictions empirically.
The following theorem provides conditions under which the L-BFGS algorithm converges superlinearly.
Its proof can be found in \cite{nocedal1999numerical}.
\begin{thm}\label{converge_lbfgs}
Let $f(\theta)$ be a twice continuously differentiable objective function. Suppose the iterates $\theta_t$ of the L-BFGS algorithm (see Sec. \ref{subsec:second_order_optim})
converge to a minimiser $\theta^*$ of $f$. Furthermore, assume that the Hessian $H$ of $f$ is Lipshitz continuous locally around $\theta^*$. Then the iterates $\theta_t$ converge superlinearly to $\theta^*$.
\end{thm}

Theorem \ref{converge_lbfgs} shows that in order to guarantee that the L-BFGS algorithm converges superlinearly to a minimum, two conditions must be checked:
\begin{enumerate}
    \item[1.] The objective function $f$ must be twice continuously differentiable.
    \item[2.] The Hessian $H$ of the objective function must be Lipshitz continuous locally about the minimum point.
\end{enumerate}

We show that ReLU/ReLU-PE-activated coordinate networks can fail to satisfy both conditions. We will first define the notion of a continuously differentiable minimum of a general continuous objective function.
\begin{defn}
Let $f$ be a continuous objective function and $\theta^*$ a (possibly local) minimum of $f$.
We say $\theta^*$ is a 
\textit{continuously differentiable (local) minimum} of $f$ if $f$ is differentiable at $\theta^*$ and the derivative is continuous at $\theta^*$. Otherwise $\theta^*$ is called a \textit{non-continuously differentiable (local) minimum}.
\end{defn}

\begin{ex}
    $ReLU(x) = \max(x,0)$, is an example of a function that contains both non-continuously differentiable and continuously differentiable minima. The point $0$ is a non-continuously differentiable minimum. This is because the derivative of ReLU is given by the function $\mathcal{H}(x) = 0$ for $x \leq 0$ and $\mathcal{H}(x) = 1$ for $x > 0$. This function is clearly not continuous at $0$. However, all negative numbers are continuously differentiable minima.
\end{ex}

\begin{ex}
    The function $f(x) = |x|$ is an example of a function with only 
    non-continuously differentiable minima, given by $x = 0$.
\end{ex}

\begin{ex}
    A sine function has only continuously differentiable minima.
\end{ex}

%\begin{enumerate}
    
%    \item[(i)] $ReLU(x) = \max(x,0)$, is an example of a function that contains both non-continuously differentiable and continuously differentiable minima. The point $0$ is a non-continuously differentiable minimum. This is because the derivative of ReLU is given by the function $\mathcal{H}(x) = 0$ for $x \leq 0$ and $\mathcal{H}(x) = 1$ for $x > 0$. This function is clearly not continuous at $0$.

%    \item[(ii)] The function $f(x) = |x|$ is an example of a function with only 
%    non-continuously differentiable minima, given by $x = 0$.

%    \item[(iii)] A sinusoidal function has only continuously differentiable minima.
%\end{enumerate}
%The ReLU function, $ReLU(x) = \max(x,0)$, is an example of a function that contains both non-continuously differentiable and continuously differentiable minima. The point $0$ is a non-continuously differentiable minimum. This is because the derivative of ReLU is given by the function $\mathcal{H}(x) = 0$ for $x \leq 0$ and $\mathcal{H}(x) = 1$ for $x > 0$. This function is clearly not continuous at $0$. On the other hand
%all negative numbers are continuously differentiable minima. The function $f(x) = |x|$ is an example of a function with only 
%non-continuously differentiable minima, given by $x = 0$.
%A sinusoidal function has only continuously differentiable minima.

\begin{ex}\label{eg:nn_loss-minima_diff}
The MSE loss function of a ReLU/ReLU-PE-coordinate network will have non-continuously differentiable minima \cite{mulayoff2021implicit}. In contrast, by the chain rule the MSE loss function of a sine/Gaussian-coordinate network 
can only have continuously differentiable minima. 
\end{ex}

%In fact, the MSE loss of a ReLU/ReLU-PE coordinate network is piecewise continuously differentiable, hence a non-continuously differentiable minimum can only occur at the knots of the function. 
%In contrast, by the chain rule the MSE loss function of a sinusoidal/Gaussian coordinate network can only have continuously differentiable minima.

%The above example highlights one of the main issues with ReLU-coordinate networks. The MSE %loss function of a ReLU-coordinate network can have non-continously differentiable minima. %This follows since a ReLU-coordinate network  

Ex.~\ref{eg:nn_loss-minima_diff} highlights a key difference between the loss landscape of a ReLU/ReLU-PE-activated network and a sine-/Gaussian-activated one, trained with MSE loss. Namely, that the former can have non-continuously differentiable minima making the Hessian about such a minimum discontinuous, while the latter will always have well behaved continuously differentiable minima. As the following theorems show, this can affect the rate of convergence of a second-order optimizer on the MSE loss of such networks.

\begin{thm}\label{relu_lbfgs_convrate}
    Let $f$ be a ReLU/ReLU-PE-activated coordinate network. Let $\mathcal{L}(\theta)$ denote the MSE loss associated to $f$ and a training set $(X, Y)$, see Sec. \ref{subsec:coordinate_nets_prelims}. 
\begin{enumerate}
    \item[1.] Then $\mathcal{L}$ is not twice continuously differentiable at every parameter point $\theta$.

    \item[2.] If the L-BFGS algorithm applied to $\mathcal{L}(\theta)$ converges to a (local) minimum $\theta^*$ such that $\theta^*$ is a non-continuously differentiable (local) minimum of $\mathcal{L}$. Then the convergence is not guaranteed to be superlinear.
\end{enumerate}
 \end{thm}

 %%%%%DO NOT DELETE THIS COMMENTED OUT PART%%%%%%%%%%%%%%%%%%%%%%%%%%
%\begin{proof}
%To prove 1, we observe that by proposition \ref{hessian relu} there will be points of parameter %space where the Hessian will contain terms consisting of step functions and Dirac delta %functions. Such terms are not continuous. By the chain rule it follows that 
%$\mathcal{L}$ is not twice continuously differentiable at every parameter point.
%
%To prove 2, we observe that at a non-continuously differentiable minimum $\theta^*$, 
%the Hessian of $\mathcal{L}$ at $\theta^*$ will not be continuous at $\theta^*$ by 
%proposition \ref{hessian relu}. By theorem \ref{converge_lbfgs}, it follows that superlinear %convergence to $\theta^*$ cannot be guaranteed. 
%\end{proof}
%%%%%DO NOT DELETE THIS COMMENTED OUT PART%%%%%%%%%%%%%%%%%%%%%%%%%%

\begin{thm}\label{sine_lbfgs_convrate}
    Let $f$ be an sine- or Gaussian-activated coordinate network. 
    Let $\mathcal{L}(\theta)$ denote the MSE loss associated to $f$ and a training set $(X, Y)$, see Sec. \ref{subsec:coordinate_nets_prelims}. 
\begin{enumerate}
    \item[1.] Then $\mathcal{L}$ is twice continuously differentiable at every parameter point $\theta$.
    \item[2.] If the L-BFGS algorithm applied to $\mathcal{L}(\theta)$ converges to a (local) minimum $\theta^*$ then the convergence is superlinear.
\end{enumerate}
 \end{thm}

Thms.~\ref{relu_lbfgs_convrate} and \ref{sine_lbfgs_convrate} show that the MSE loss of a sine- or Gaussian-coordinate network 
has continuous curvature across parameter space, while ReLU-activated networks do not. This makes second-order optimizers effective in accelerating the training of sine-/Gaussian-activated networks, compared to first-order optimizers such as SGD or Adam, which generally have sub-linear/linear rates of convergence \cite{garrigos2023handbook, kingma2014adam}.

%Thms.~\ref{relu_lbfgs_convrate} and \ref{sine_lbfgs_convrate} demonstrate a crucial distinction between the curvature of the MSE loss of sinusoidal/Gaussian-activated and ReLU-activated coordinate networks. While the former has continuous curvature across the parameter space, the latter does not. As a result, using curvature-aware second-order optimizers can significantly accelerate the training process for sinusoidal/Gaussian-activated coordinate networks in comparison to first order optimizers such as SGD or Adam, which in general have 
%sub-linear/linear rates of convergence \cite{garrigos2023handbook, kingma2014adam}.

%%%%%DO NOT DELETE THIS COMMENTED OUT PART%%%%%%%%%%%%%%%%%%%%%%%%%%
%\begin{proof}
%To prove 1, we simply observe that a sine or Gaussian functions is twice continuously %differentiable. This implies that a sine- or Gaussian-activated coordinate network is twice %continuously differentiable via the chain rule. Another application of the chain rule, see the %definition of $\mathcal{L}$ in \eqref{loss}, implies 1 holds.
%
%To prove 2, we first observe that since the second derivatives of a sine or Gaussian function %is Lipshitz continuous, the chain rule, and the formula for $\mathcal{L}$ given in %\eqref{loss}, implies that the Hessian $H$ of $\mathcal{L}$ is Lipshitz continuous about a %neighbourhood of $\theta^*$. 
%Applying theorem \ref{converge_lbfgs} we find that the convergence rate is superlinear.
%\end{proof}
%%%%%DO NOT DELETE THIS COMMENTED OUT PART%%%%%%%%%%%%%%%%%%%%%%%%%%

Fig. \ref{fig:fig_1} shows the convergence of a sine and ReLU-activated coordinate network trained with both Adam and L-BFGS on an image reconstruction task. The sine-network trained with L-BFGS has a much faster convergence rate than the sine-activated network trained with Adam, as predicted by Thm.~\ref{sine_lbfgs_convrate}. However, the ReLU-network trained with Adam converges at a faster rate than L-BFGS, see Thm~\ref{relu_lbfgs_convrate}.

%\begin{figure}[t]
%    \includegraphics[width=0.9\linewidth]{figures/image/ICCV_LBFGS_Adam_ReLU_Pepper.pdf}
%    \vspace{-1em}
%    \caption{The ratio of convergence time,$\frac{{t^c}_{Adam}}{{t^c}_{L-BFGS}}$, where $t^c$ denotes the time taken for convergence, for optimizing a ReLU-activated MLP using L-BFGS and Adam optimizer on a pepper image. A value less than 1 indicates Adam converged faster. Note that both lines are under the {\textcolor{red}{red}} line. }
%    \label{fig:lbfgs_vs_adam_relu}
%\end{figure}

%\subsection{L-BFGS: An empirical analysis}

%\begin{figure}
 %   \centering
 %   \hspace{-0.5cm}
 %   \includegraphics[width=0.5\textwidth]{figures/image/eigenvals.pdf}
 %   \caption{Caption}
%    \label{fig:eigenvals}
%\end{figure}
\begin{figure*}[t]
    \centering
    \includegraphics[width=1\linewidth]{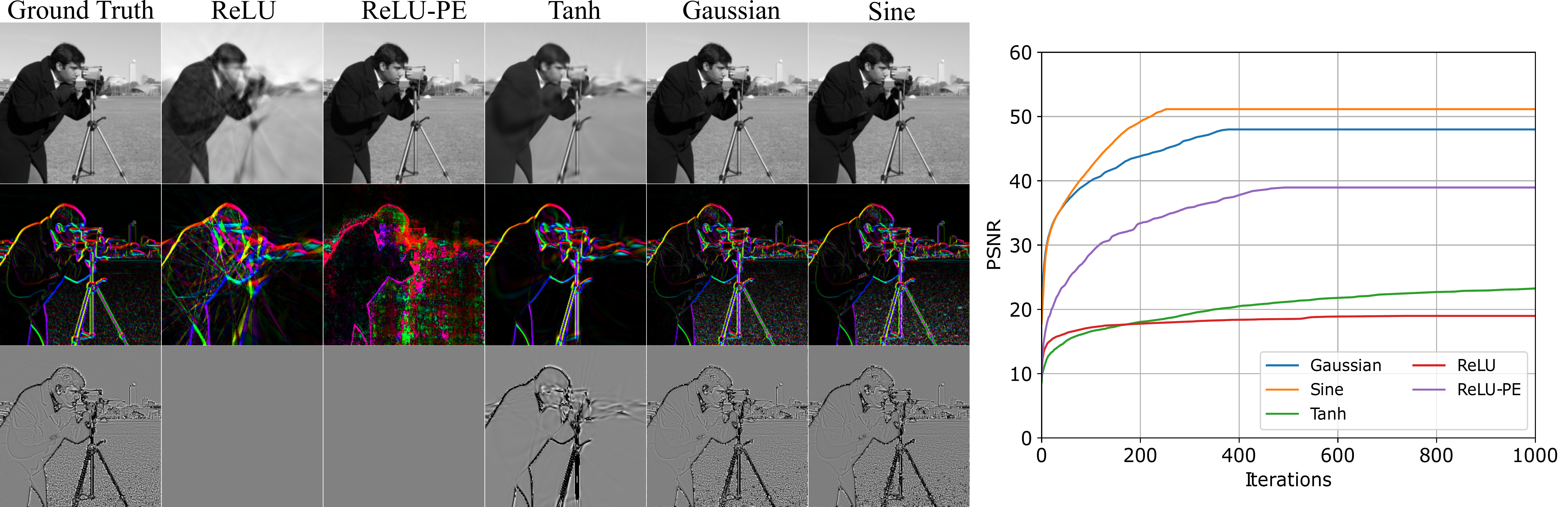}
    \vspace{-1em}
    \caption{ \textbf{2D Image Reconstruction.}  \textit{Left}: Comparison of various coordinate-MLPs $f$ in fitting the \textit{Cameraman} image (\textit{top left}) using the L-BFGS optimizer. Note that all networks were only trained on the target image. We also show gradient (\emph{second row}) and Laplacian (\emph{third
    row}) of neural output function. \textit{Right}: Training convergence of each network.}
    \label{fig:gradient_image}
\end{figure*}

\section{Experiments}\label{sec:Exps}
In this section, we demonstrate the effectiveness of L-BFGS on various popular tasks: 2D image reconstruction and novel view synthesis using neural radiance fields (NeRF); see Sec. 4 of supp. material for additional results for other modalities such as audio, shape and video reconstruction.
%We used PyTorch's implementation of L-BFGS, which requires three main hyperparamters, a learning rate, history size and max iterations. The history size is the number of previous curvature terms stored by the algorithm in order to compute the new Hessian approximation at a particular iterate.
%Doing a parameter sweep, we found that a learning rate = 1, history size = 5 and the max iterations = 40, provided the best results. 

\begin{figure}[b]\centering
    \includegraphics[width=0.9\linewidth]{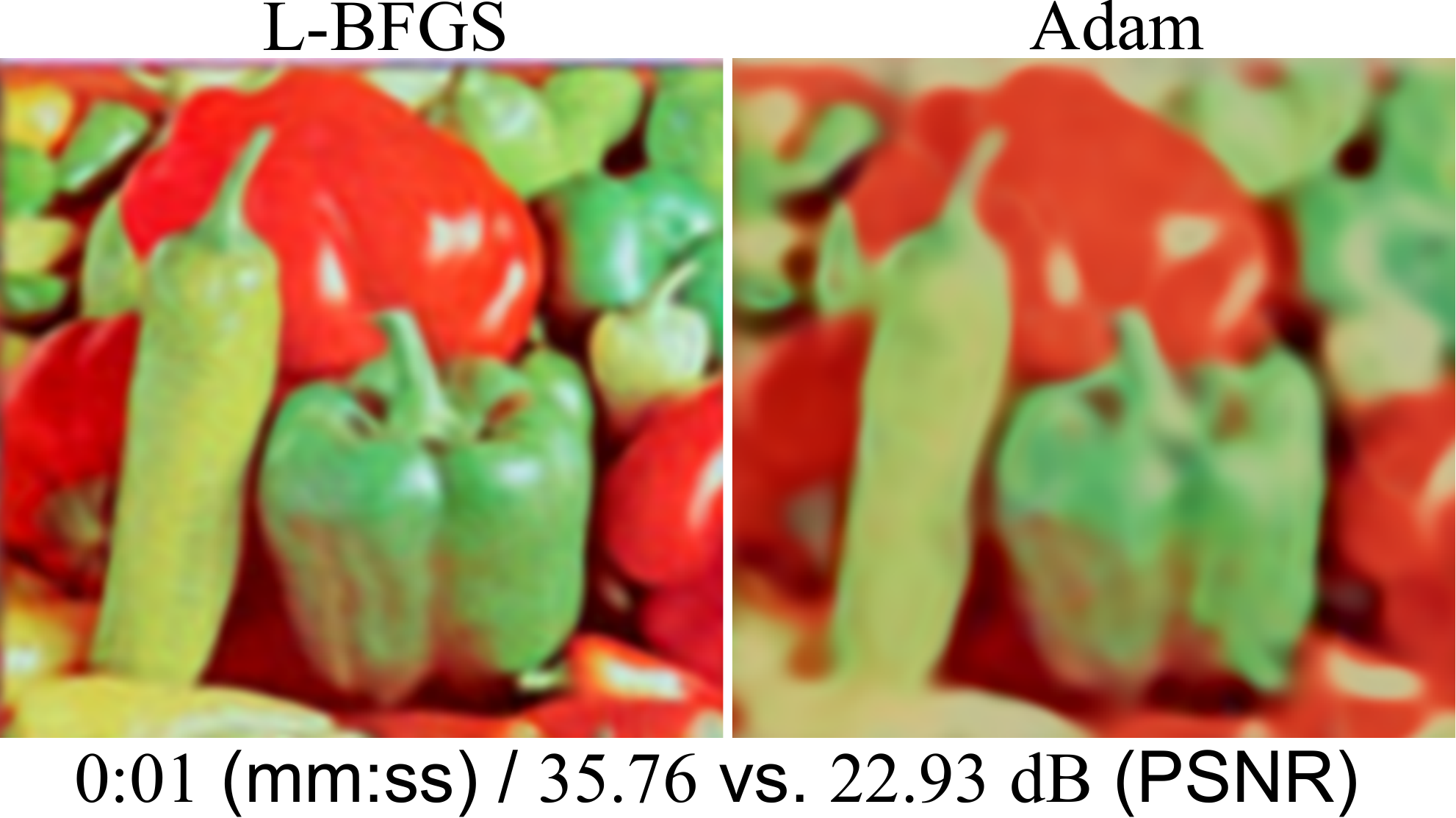}
     \vspace{-0.5em}
    \caption{\textbf{2D Image Reconstruction.} L-BFGS has achieved a substantially better reconstruction than Adam given the same amount of training time. }%Comparison of L-BFGS and Adam on optimizing a sine-activated MLP, using data set sizes from 5-360k on an upsampled 600x600 peppers image.}
    \label{fig:lbfgs_vs_adam_sine}
\end{figure}

\begin{figure}[b]\centering
    \includegraphics[width=0.92\linewidth]{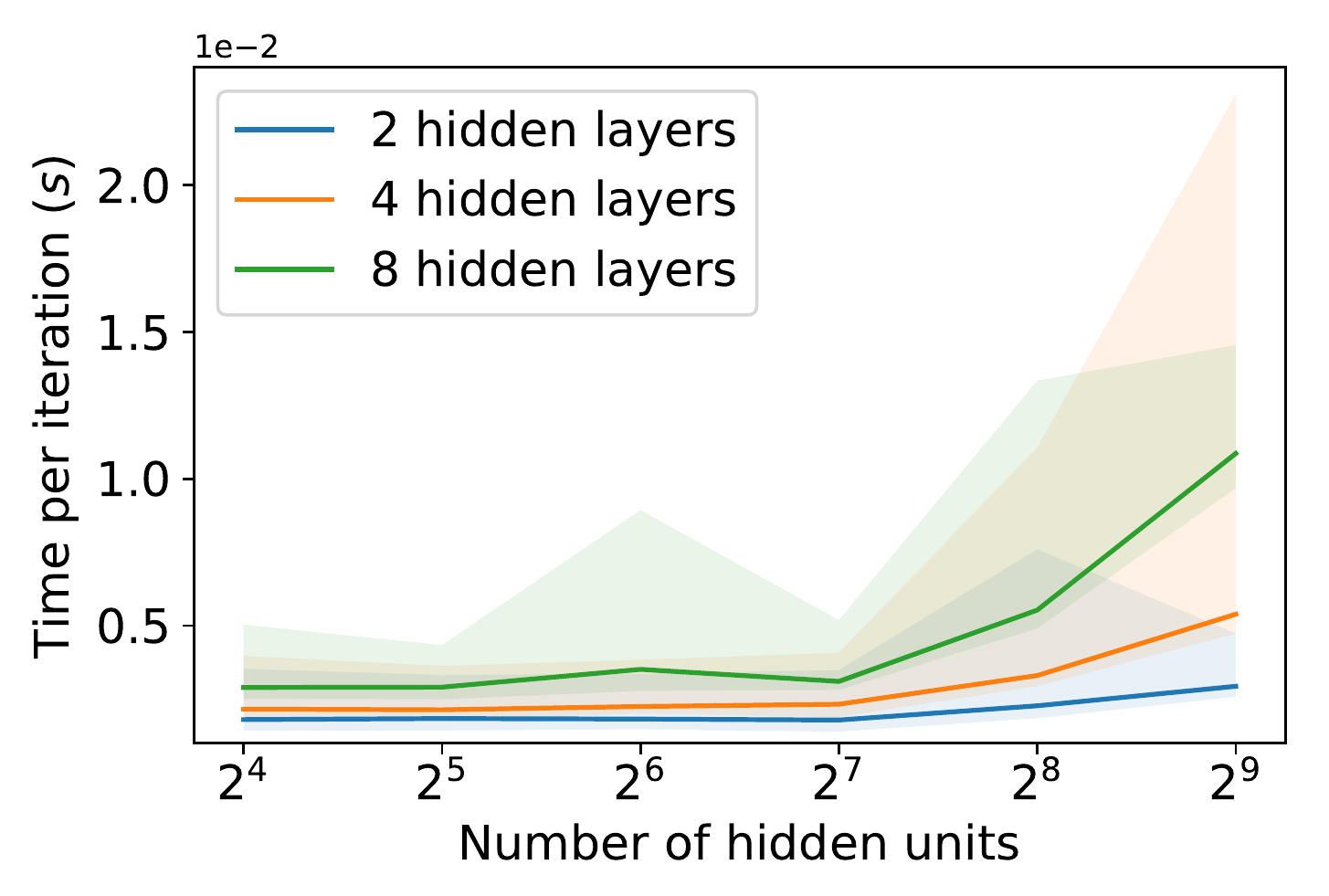}
     \vspace{-1em}
    \caption{As the size of the network parameters grow, the time per iteration for the L-BFGS optimizer increases due to the added computational complexity. Note that \textit{solid} line denotes mean while \textit{transparency region} denotes variance.}
    \label{fig:lbfgs_computational}
\end{figure}

\begin{figure*}[t]\centering
    \includegraphics[width=1\linewidth]{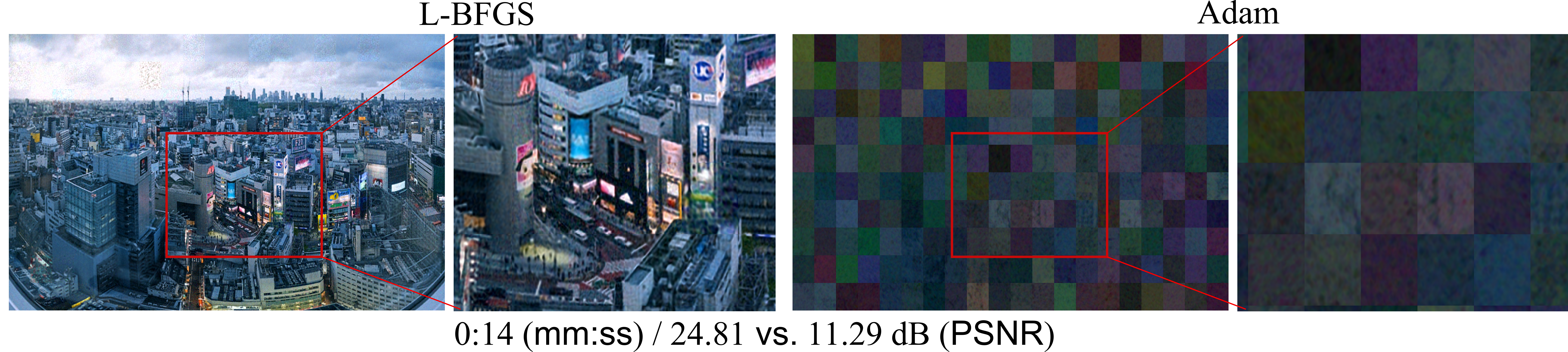}
     \vspace{-2em}
    \caption{\textbf{Gigapixel Image Reconstruction.} Our approach represents a gigapixel image using $200$ sine-activated tiny-MLPs. We report a comparison in terms of optimization time and PSNR (L-BFGS vs. Adam). Using L-BFGS, our method achieves a substantially higher-fidelity reconstruction (\textbf{24.81} dB) than Adam (11.29 dB) given the same amount of training time. L-BFGS achieves convergence $\sim 14\times$ faster than Adam ($33.8$ vs. $33.22$ dB).}
    \label{fig:kilo_image}
\end{figure*}

\subsection{Images}\label{subsec:exp_image}
Given pixel coordinates $\mathbf{x} \in \mathbb{R}^2$, we aim to optimize the network $f$ to regress the associated RGB values $\mathbf{c} \in \mathbb{R}^3$~\cite{siren, ramasinghe22}. In this task, we will first explain why non-traditional activations are well-suited for training with L-BFGS, followed by their comparisons with that of competitive first-order optimizers, e.g. Adam.
%An RGB image $g$ defines a discrete data set consisting of points $\{(\textbf{z}_i, g(\textbf{z}_i)\}$, where $\textbf{z}_i = (x_i, y_i)$ denotes pixel coordinates and $g(\textbf{z}_i)$ denotes the associated RGB value. We seek to find a continuous map $f : \R^2 \rightarrow \R^3$ that parameterises the discrete image $g$. A standard approach in reconstructing the image $g$ is to use a coordinate-MLP to learn the continuous function $f$ \cite{siren, ramasinghe22}. 
\vspace{-1em}
\paragraph{Gradient Perspective.} 
In Fig.~\ref{fig:gradient_image}, we compare the performance of various network architectures optimized with L-BFGS on the Cameraman image using a 4-hidden layer MLP with 64 hidden units. 
As predicted by Prop.~\ref{hessian relu} and Thm.~\ref{relu_lbfgs_convrate}, ReLU and ReLU-PE activations produce \textit{extremely bad} gradients and Laplacian, while Tanh lacks fine details. In contrast, sine and Gaussian activations produce high-quality reconstructions with \textit{well-behaved derivatives}. Furthermore, non-traditional activations converge \textit{significantly faster} than traditional ones, indicating that traditional activations do not train well with L-BFGS, as predicted by Thm.~\ref{sine_lbfgs_convrate}.

\vspace{-1em}
\paragraph{L-BFGS vs Adam.} 
Fig.~\ref{fig:lbfgs_vs_adam_sine} shows a reconstruction snapshot of the \textit{pepper} image, trained with both L-BFGS and Adam, using a 4-hidden layer, 64 width sine-activated network. Impressively, despite L-BFGS only being trained for $1s$, the reconstruction is remarkably sharp ($35.76$dB), compared to Adam which achieved a PSNR of $22.93$. Overall, L-BFGS achieves convergence $5\times$ faster than Adam.

 \vspace{-1em}
\paragraph{Computational Bottleneck.}\label{para:computational_bottleneck}
We found that when training with a large size neural network, L-BFGS did not offer any significant advantage over Adam. Fig.~\ref{fig:lbfgs_computational} shows that the computational time of L-BFGS increases when the network's parameter size grows, 
due to its computational complexity for computations using past curvature vectors, see Sec.~\ref{subsec:second_order_optim}. To mitigate this issue, we propose a patch-based decomposition strategy in Sec. \ref{sec:kilo_image}.

\subsection{KiloImage}\label{sec:kilo_image}
We introduce KiloImage, a patch-based strategy for optimizing a gigapixel image using L-BFGS. KiloImage decomposes the image into small grids, each represented via a small sine-MLP with $4$ hidden layers and $64$ neurons. We sample 10k points on each grid and combine the resulting outputs to form a global reconstruction. Fig.~\ref{fig:kilo_image} shows that L-BFGS outperforms Adam, achieving on average $\sim14$ times faster convergence, with high-quality reconstructions after just $0.14$ seconds of training. For additional results on other gigapixel instances, please refer to Sec. 4 of the supp. material.

%shows that L-BFGS outperforms Adam, achieving an overall of convergence $\sim14$ times faster with high-quality reconstructions

\begin{figure}[b]
    \centering
    %\begin{subfigure}[b]{0.5\textwidth}
    %\includegraphics[width=\textwidth]%{figures/image/tiny_nerf_150secs.pdf} 
   % \caption{Trained for 150s}
   % \end{subfigure}
    %\begin{subfigure}[b]{0.5\textwidth}
    \includegraphics[width=0.48\textwidth]{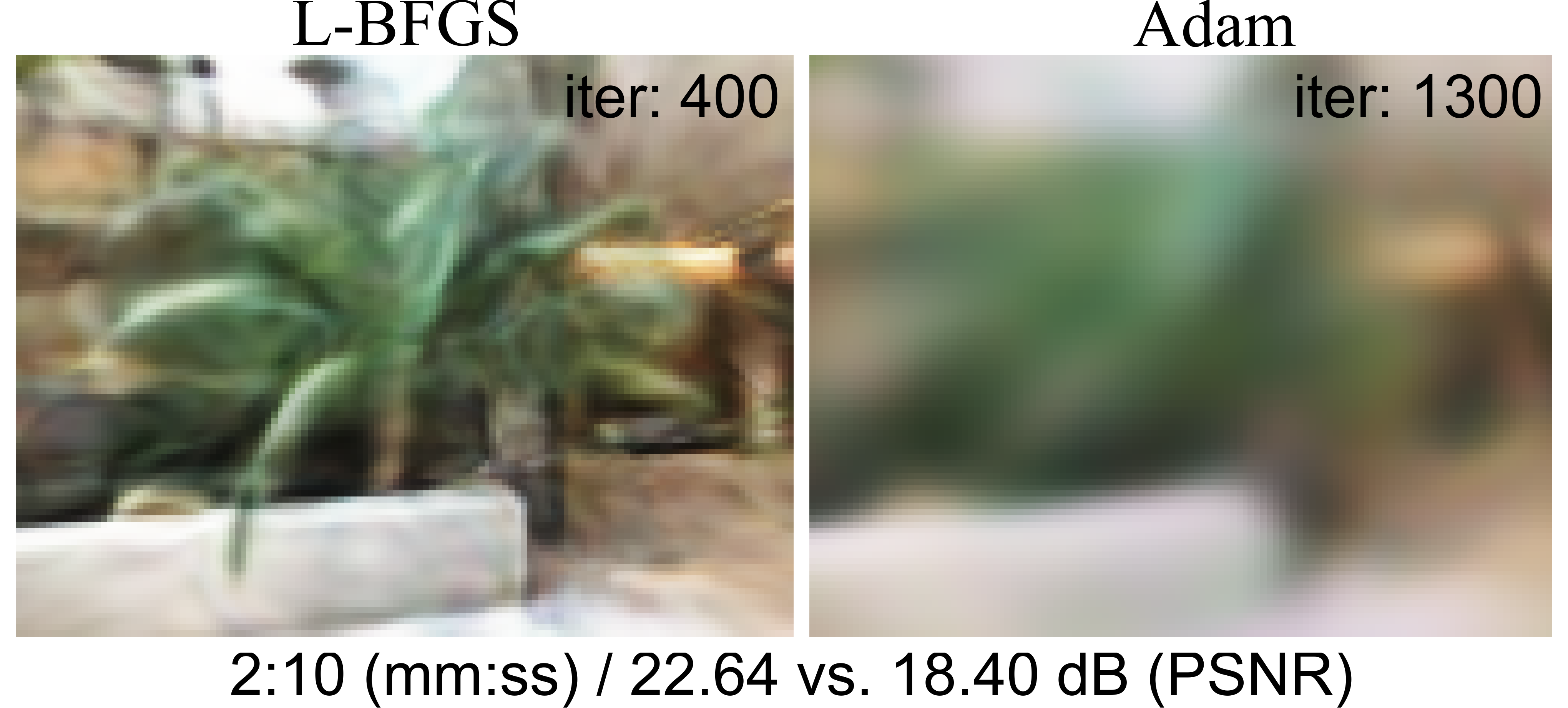}
    %\caption{Trained for 350s}
    %\end{subfigure}
    \vspace{-1em}
    \caption{\textbf{Novel View Synthesis with NeRF} for a \emph{fern} instance from the LLFF dataset~\cite{mildenhall2019local}. We report a comparison in terms of optimization time and PSNR (L-BFGS vs. Adam). Using L-BFGS, Tiny-NeRF achieves a good reconstruction (\textbf{22.64}dB) compared to Adam (18.40dB) after training for the same amount of time. Overall, L-BFGS achieves convergence ($24$dB) $\sim 2\times$ faster than Adam.}
    \label{fig:tiny_nerf}
\end{figure}

\subsection{Neural Radiance Fields (NeRF)}\label{sec:nerf}
NeRF has recently emerged as a compelling strategy for utilizing a MLP to model 3D objects and scenes using multi-view 2D images. This approach shows promise for generating high-fidelity reconstruction in novel view synthesis task~\cite{nerf,kilonerf,chng2022garf,lin2021barf}.
Given 3D points $\mathbf{x} \in \mathbb{R}^3$ and viewing direction, NeRF aims to estimate the radiance field of a 3D scene which maps each input 3D coordinate to its corresponding volume density $\sigma \in \mathbb{R}$ and directional emitted color $\mathbf{c} \in \mathbb{R}^3$~\cite{nerf,lin2021barf,chng2022garf}. 
In this section, we compare L-BFGS against the Adam optimizer on a popular application of a coordinate network in novel view synthesis task, NeRF~\cite{nerf}. 
For simplicity, we used a minimalist version of a NeRF model that excluded view-dependence and hierarchical ray sampling. We trained a tiny Gaussian-activated MLP with 4 hidden layers and 128 neurons on the real world LLFF forward-facing scenes~\cite{nerf}, that were downscaled by a factor of 5. 
Fig.~\ref{fig:tiny_nerf} presents the qualitative result obtained for the 
\emph{fern} instance. Impressively, with only $130$ seconds, TinyNeRF trained with L-BFGS generated a detailed reconstruction in only $400$ iterations whereas TinyNeRF trained with Adam produced blurry renderings for $1300$ iterations, indicating the superiority of L-BFGS for this instance.

\begin{figure}
    \centering
    \includegraphics[width=0.46\textwidth]{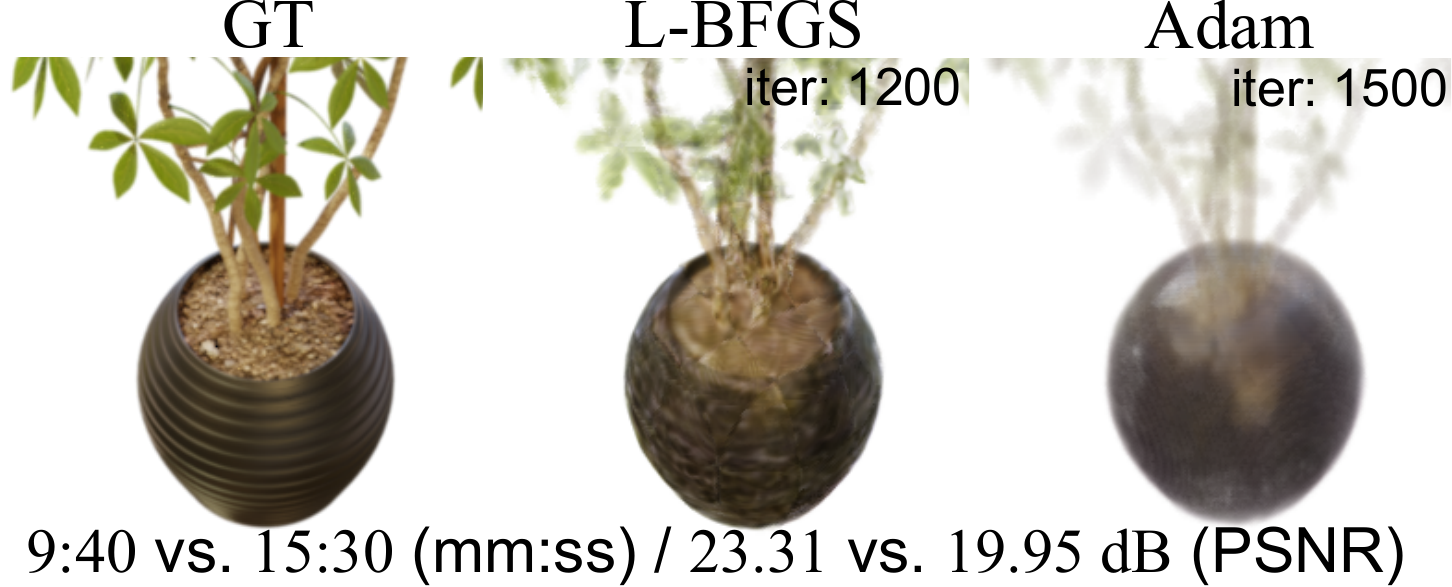}
    %\caption{Trained for 350s}
    %\end{subfigure}
    %\vspace{-0.5em}
    \caption{\textbf{Novel View Synthesis with KiloNeRF} for a \emph{ficus} test instance from the LLFF dataset~\cite{mildenhall2019local}. We report a comparison in terms of optimization time and PSNR (L-BFGS vs. Adam). Using L-BFGS, KiloNeRF achieves a good reconstruction (\textbf{23.31}dB) compared to Adam (19.95dB).  }
    \label{fig:kilonerf_ficus}
\end{figure}

\vspace{-1em}
\paragraph{KiloNeRF.}
As discussed in Sec.~\ref{subsec:exp_image}, while L-BFGS can achieve faster convergence compared to Adam, this advantage diminishes as the number of parameters of the neural network increases. This presents a particular challenge when training NeRF, a 5D high-dimensional problem that typically requires a larger network, such as an 8-layer 256 network. In this section, we showcase how we can mitigate the computational bottleneck associated with training a large-scale NeRF with L-BFGS. Building on the recent innovation of KiloNeRF \cite{kilonerf}, we trained thousands of 2-layer 32 width sine-activated KiloNeRF with L-BFGS and compared its performance to one trained with Adam. Note that we sampled each MLP with 10k points. As presented in Table~\ref{tab:kilonerf}, L-BFGS trained the KiloNeRF $\sim 6\times$ faster than Adam and produced higher reconstruction quality along the way. Fig. \ref{fig:kilonerf_ficus} shows a qualitative result of a KiloNeRF trained with L-BFGS and Adam, after $600$ seconds the L-BFGS trained KiloNeRF is already able to reconstruct with good quality; see Sec. 4 of supp. material for more qualitative results.
%Considering only a 10K fixed sampling strategy was used, this paves the way of using such techniques with L-BFGS to obtain significant speed-ups in training time.

%this advantage diminishes as the number of parameters of neural network increase.

%we trained thousands of 2-layer 32 sine-activated KiloNeRF

\begin{table}[t]
    \setlength{\tabcolsep}{4.5pt}
    \renewcommand{\arraystretch}{1}
	\centering
    \begin{tabular}{lccccc}
        \toprule
        Method & PSNR$\uparrow$  & SSIM$\uparrow$ & LPIPS$\downarrow$ & Num & Training  \\
         &  &  &  &  Iters$\downarrow$ & Time ($s$)$\downarrow$  \\
        \midrule
        Adam & $ 20.78$  & $0.85$ & $0.18$ & 6000 & 5324.48\\
        L-BFGS & $\textbf{22.01}$  & $\textbf{0.86}$ & $\textbf{0.15}$ & \textbf{1200} & $\textbf{889.97}$\\
        \bottomrule
	\end{tabular}
	%\vspace{-1em}
	\caption{Quantitative results of KiloNeRF on all instances from the Blender dataset~\cite{nerf}. On average, L-BFGS trained $6\times$ faster than Adam and was able to achieve competitive quality scores with significantly less iterations.}
	% \vspace{-0.3cm}
 \label{tab:kilonerf}
\end{table}

%after 350s its reconstruction is much sharper. Considering only a 10K fixed sampling strategy was used, this paves the way of using such techniques with L-BFGS to obtain significant speed ups in training time.

%\begin{minipage}\includegraphics[width=0.5\textwidth]{figures/image/kilonerf_100s.pdf}\end{minipage}

%\begin{figure}
%    \centering
%    \begin{subfigure}[b]{0.5\textwidth}
%    \includegraphics[width=\textwidth]{figures/image/kilonerf_100s.pdf} 
%    \caption{Rendered images using KiloNeRF trained for 150s, and  finetuned for 1K iterations.}
%    \end{subfigure}
%    \begin{subfigure}[b]{0.5\textwidth}
%    \includegraphics[width=\textwidth]{figures/image/kilonerf_200s.pdf}
%    \caption{Rendered images using KiloNeRF trained for 350s, and  finetuned for 1K iterations}
%    \end{subfigure}
%    \vspace{-1em}
%    \caption{Rendered image using KiloNeRF. \emph{Left}: GroundTruth. \emph{Centre}: L-BFGS. \emph{Right}: Adam}\label{fig:kilo_nerf}
%\end{figure}

\begin{figure}
    \centering
    \begin{subfigure}{0.47\textwidth}
        \includegraphics[width=\textwidth]{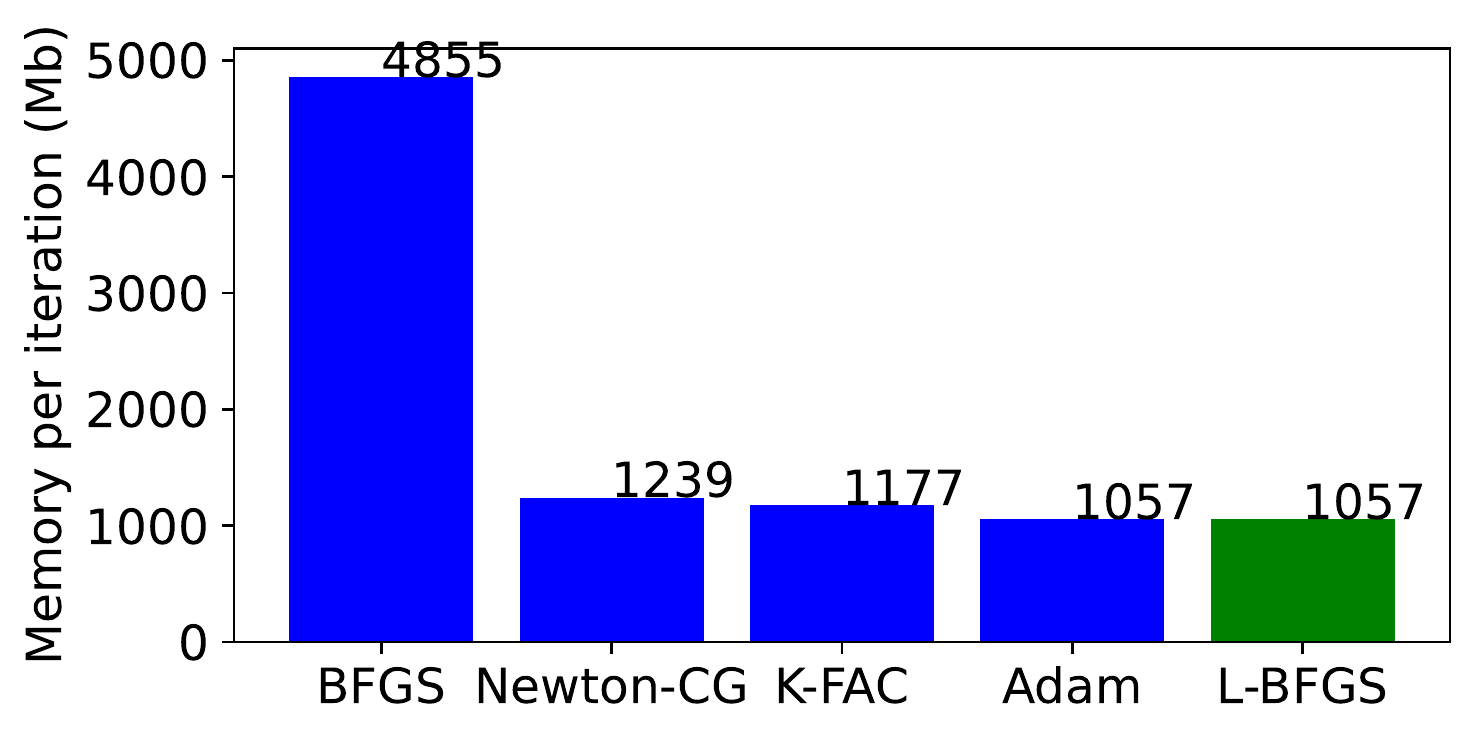}
        \vspace{-2em}
        \caption{Memory per iteration}
    \end{subfigure}
    \begin{subfigure}{0.47\textwidth}
        \includegraphics[width=\textwidth]{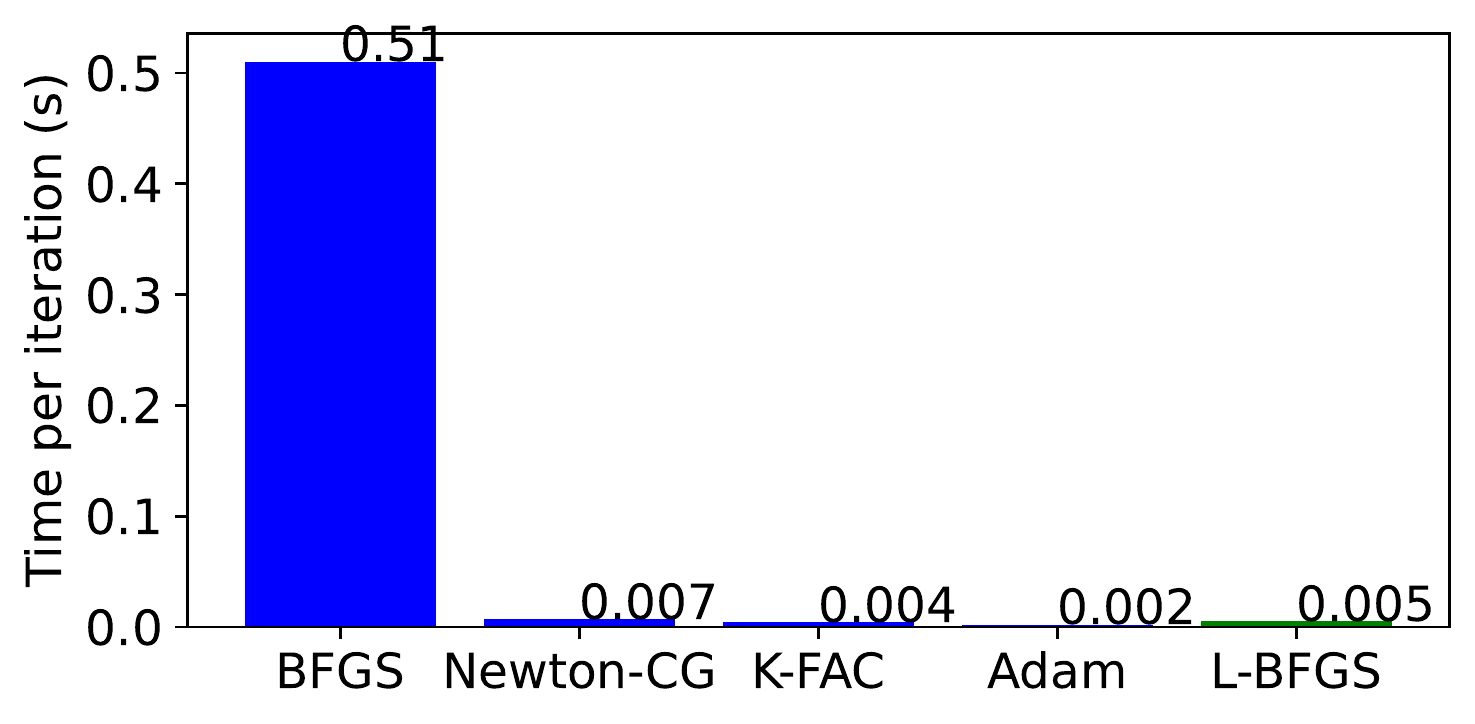}
        \vspace{-2em}
        \caption{Training time per iteration}
    \end{subfigure}
    %\caption{Trained for 350s}
    %\end{subfigure}
    \vspace{-0.5em}
    \caption{Comparison of computational complexity of various well-known second-order optimizers. L-BFGS strikes a good balance between fast training and comparable memory usage compared to Adam.}
    \label{fig:memory}
\end{figure}

%\subsection{Memory Complexity of L-BFGS}
%\begin{itemize}
%    \item history size versus memory
%    \item number of parameters versus memory (redundant: since it affects adam as well)
%    \item compare again BFGS too 
%\end{itemize}

%\subsection{3D Shape Regression}

%Here, we compare L-BFGS against Adam on a 3D shape regression function. In this task, we aim 
%to learn a binary occupancy field, which represents a 3D shape as the decision boundary of an MLP~\cite{wang2021spline, gropp2020implicit, atzmon2020sal}.  Our setup was a 4 layer, 64 neuron, sinusoidal activated coordinate-MLP. Samples were drawn from a
%100k point cloud. Figure \ref{fig:sdf} shows that when trained with L-BFGS, a complete reconstruction is possible after only just $3.5s$, while at this point Adam is still unable to produce a full reconstruction.

%\begin{figure}
%    \centering
%     \includegraphics[width=1.0\linewidth]{figures/image/tiny_sdf.pdf}
%     \hspace{-0.2cm}
%    \caption{Comparisons on reconstructed mesh. \emph{Left}: GroundTruth. \emph{Center}: L-BFGS. \emph{Right}: Adam.}\label{fig:sdf}
%\end{figure}

\subsection{Analyzing the Computational Cost}
Fig. \ref{fig:memory} presents a comparison of different second-order optimizers in terms of memory usage and optimization time per iteration, see Sec. 1 of supp. material for details on these optimizers.
Interestingly, we found that K-FAC \cite{martens2015optimizing} struggled to optimize a sine-activated coordinate network. This led us to speculate that the initialization scheme proposed in~\cite{siren} may not be optimal for K-FAC optimization and warrants further investigation. The results in Fig. \ref{fig:memory} were obtained using a 4-layer 64 width Gaussian-activated coordinate-MLP. Our findings indicate that L-BFGS strikes a good balance between fast training and comparable memory usage compared to Adam, offering a ``best of both worlds" solution. While Adam has the lowest time and memory per iteration, it is important to note that second-order optimizers aim to converge with fewer iterations. This is demonstrated in the experiments discussed in Sec. \ref{sec:Exps}.

%Fig.~\ref{fig:memory} compares the computational cost of different well-known second-order optimizers in terms of memory usage and optimization time per iteration. Interestingly, we observed that K-FAC was not able to optimize a sine-activated coordinate network. Based on this, we postulated that the initialization scheme proposed in~\cite{siren} may no longer be optimal for K-FAC optimization; we leave this as a future work. The results presented in 
%Fig.~\ref{fig:memory} were obtained for a Gaussian activated coordiante network. Overall, our results suggest that L-BFGS offers a "best out of two worlds" solution with fast training convergence and comparable memory usage compared to Adam.

%Note that the results presented in Fig.~\ref{fig:memory} for K-FAC were obtained using a Gaussian-activated coordinate network. Overall, our results suggest that L-BFGS offers a "best out of two worlds" solution with fast training convergence and comparable memory usage compared to Adam.

\section{Conclusion}
This paper analyzes the use of second-order optimizers in training coordinate-MLPs. We show that coordinate networks using non-traditional activations, such as sine or Gaussian functions, have better conditioned Hessians/curvature than ReLU-coordinate networks. We tested our theory using an L-BFGS optimizer on a range of signal reconstruction tasks and found that using second-order optimizers on small-scale data projects significantly reduces training times compared to using Adam. However, for large-scale applications with a large number of parameters, the computational complexity of a second-order optimizer hinders its use. To address this issue, we proposed a patch-based decomposition of large datasets into smaller patches, training a second-order optimizer on each patch, providing a viable solution for training a coordinate network with a second-order optimizer on large-scale applications.
\vspace{-0.5em}
\section{Limitations}
%\vspace{-0.5em}
%The training with L-BFGS on all the coordinate-MLPs in the experiments in this paper were done with full samples. 
%A limitation of this work is that we couldn't get any form of stochastic sampling working for L-BFGS. While there have been  specific variants of L-BFGS that allow for stochastic sampling 
%\cite{zhao2017stochastic}, \cite{stochastic_lbfgs1}, \cite{stochastic_lbfgs2}
%we found that when trained on a coordinate network the algorithm performed poorly in comparison to an Adam optimizer trained with stochastic sampling. Thus it remains a challenge for practitioners in the field of coordinate-MLPs to understand how stochastic second-order optimizers can be used to train coordinate-MLPs.

%The experiments in this paper trained all coordinate-MLPs using full samples with L-BFGS. However, a limitation of this work is the inability to implement stochastic sampling with L-BFGS. Although some variants of L-BFGS allow for stochastic sampling, such as those discussed in \cite{zhao2017stochastic}, \cite{stochastic_lbfgs1}, and \cite{stochastic_lbfgs2}, our experiments show that when used to train coordinate networks, L-BFGS performs poorly compared to an Adam optimizer that employs stochastic sampling. This presents a challenge for practitioners in the field of coordinate-MLPs who wish to utilize stochastic second-order optimizers in their training.

We trained coordinate-MLPs using L-BFGS with full samples. While stochastic L-BFGS exists (e.g., \cite{zhao2017stochastic, stochastic_lbfgs1, moritz2016linearly}), we found that such stochastic L-BFGS algorithms perform poorly for training coordinate-MLPs compared to Adam with stochastic sampling. This presents a challenge for practitioners who want to use stochastic second-order optimizers in training coordinate-MLPs.

% Update the cvpr.cls to do the following automatically.
% For this citation style, keep multiple citations in numerical (not
% chronological) order, so prefer \cite{Alpher03,Alpher02,Authors14} to
% \cite{Alpher02,Alpher03,Authors14}.

%------------------------------------------------------------------------

%------------------------------------------------------------------------
%\section{Final copy}

%You must include your signed IEEE copyright release form when you submit your finished paper.
%We MUST have this form before your paper can be published in the proceedings.

%Please direct any questions to the production editor in charge of these proceedings at the IEEE %Computer Society Press:
%\url{https://www.computer.org/about/contact}.
\clearpage

{\small
\bibliographystyle{ieee_fullname}
\bibliography{egbib}
}

\end{document}